%% file: colm2025_conference.tex
\documentclass{article}
\usepackage[submission]{colm2025_conference}

\usepackage{microtype}
\usepackage{hyperref}
\usepackage{url}
\usepackage{booktabs}

\usepackage{lineno}
\usepackage{graphicx}
\usepackage{xspace}
\usepackage{longtable}
\usepackage{wrapfig}

\definecolor{darkblue}{rgb}{0, 0, 0.5}
\hypersetup{colorlinks=true, citecolor=darkblue, linkcolor=darkblue, urlcolor=darkblue}

\title{Evaluation Under Imperfect Benchmarks and Ratings:\\ A Case Study in Text Simplification}

\author{Joseph Liu, Yoonsoo Nam, Xinyue Cui \& Swabha Swayamdipta \\
Department of Computer Science\\
University of Southern California \\
\texttt{\{jliu7350, yoonsoon, xinyuecu, swabhas\}@usc.edu} 
}

\colmpreprinttrue

\newcommand{\benchmarkname}{{\fontfamily{lmss}\selectfont{SynthSimpliEval}}\xspace}
\newcommand{\method}{{\fontfamily{lmss}\selectfont{LLMs-as-a-Jury}}\xspace}

\begin{document}

\maketitle

\begin{abstract}

Despite the successes of language models, their evaluation remains a daunting challenge for new and existing tasks.
We consider the task of text simplification, commonly used to improve information accessibility, where evaluation faces two major challenges.
First, the data in existing benchmarks might not reflect the capabilities of current language models on the task, often containing disfluent, incoherent, or simplistic examples.
Second, existing human ratings associated with the benchmarks often contain a high degree of disagreement, resulting in inconsistent ratings; nevertheless, existing metrics still have to show higher correlations with these imperfect ratings.
As a result, evaluation for the task is not reliable and does not reflect expected trends (e.g., more powerful models being assigned higher scores).
We address these challenges for the task of text simplification through three contributions. 
First, we introduce \benchmarkname, a synthetic benchmark for text simplification featuring simplified sentences generated by models of varying sizes.
Through a pilot study, we show that human ratings on our benchmark exhibit high inter-annotator agreement and reflect the expected trend: larger models produce higher-quality simplifications. 
Second, we show that auto-evaluation with a panel of LLM judges (\method) often suffices to obtain consistent ratings for the evaluation of text simplification.
Third, we demonstrate that existing learnable metrics for text simplification benefit from training on our \method-rated synthetic data, closing the gap with pure \method for evaluation.
Overall, through our case study on text simplification, we show that a reliable evaluation requires higher quality test data, which could be obtained through synthetic data and \method ratings.\footnote{Code can be found at \hyperlink{https://github.com/jliu7350/text-simplification-benchmark}{https://github.com/jliu7350/text-simplification-benchmark}}

\end{abstract}

\section{Introduction}
\label{sec:intro}

Despite advances in LLMs, evaluating the quality of their generations remains a challenge \cite{pillutla2021mauvemeasuringgapneural, chang2023surveyevaluationlargelanguage}.
One such task is text simplification: crucial to improving the clarity and accessibility of information, making content easier to understand for a wider audience \citep{al2021automated}. 
The gold standard for text simplification evaluation is human judgment, which provides the most direct and comprehensive assessment of simplification quality \citep{devaraj-etal-2022-evaluating, maddela-etal-2023-lens}.
However, existing human evaluation of text simplification can be unreliable due to low agreement among annotators on simplification ratings \citep{wu2024depth, popovic-etal-2022-reproducing}.
This makes it difficult to establish a universally reliable evaluation standard.

To address this challenge, we propose a novel text simplification benchmark named \benchmarkname featuring a dataset of complex sentences and their corresponding simplifications generated by LLMs of varying sizes.
Through a pilot study with new human annotators, we observe high inter-annotator agreement on our benchmark.
Moreover, the human study  shows a strong correlation between the score assigned to each simplification and the size of the model that generated it. 
This finding is consistent with prior research demonstrating that, within the same model family, larger models consistently outperform smaller ones across a range of NLP tasks \citep{hestness2017deep, kaplan2020scalinglawsneurallanguage, hoffmann2022training, liang2023holisticevaluationlanguagemodels, mckenzie2023inverse}.

To alleviate the cost of human labeling, we subsequently adopt an \method approach \citep{verga2024replacing, chan2023chateval, wang2024mixture} to evaluate simplification quality, aggregating the scores across all models in the jury to obtain the final score. 
We conduct a systematic ablations study on how various design choices, such as prompting strategy, rationale inclusion, and aggregation method, affect the assigned simplification score. 
We find that few-shot prompting with rationale generation, combined with score averaging, yields simplification scores that best correlate with model sizes.
Moreover, we compare inter-LLM agreement for unified scoring against the multi-dimensional approach used in previous benchmarks \citep{wubben-etal-2012-sentence, maddela-etal-2023-lens, alva-manchego-etal-2020-asset}, and find that unified scoring results in higher agreement. This suggests that unified scoring is both simpler to apply and more reliable as an evaluation method.

Motivated by the observed trend that larger models produce higher-quality simplifications, we evaluate the performance of existing text simplification metrics \citep{flesch1948new, maddela-etal-2023-lens, zhang2019bertscore, cripwell2023simplicity} and \method on \benchmarkname by measuring the correlation between assigned scores and model sizes. Existing metrics struggle to consistently reflect the  trend, whereas \method reliably assigns higher scores to outputs from larger models, aligning with expectations on simplification quality.

In order to improve existing learnable metrics using our synthetic data, we use the same approach to generate synthetic complex sentences and create a set of LLM-rated complex-simple sentence pairs.
We train a small neural network on sentence embeddings of these scores, following previous work \citep{maddela-etal-2023-lens, huang-kochmar-2024-referee}.
We find that with this new set of training data, the network recovers some of the capability of the LLMs, achieving a higher correlation with model size than all previous methods.

Overall, through our case study on text simplification, we show that a reliable evaluation requires high-quality data.
By leveraging synthetic benchmarks in place of traditional human annotations, we are able to reliably evaluate text simplification metrics. 
This approach offers a practical recipe for evaluation in other tasks where high-quality annotated data may be limited or unreliable.
We will publicly release our code and data.

\section{Existing Text Simplification Benchmarks and Ratings}
\label{sec:existing_benchmarks_ratings}

When evaluating text simplification metrics, existing work often relies on datasets that contain both complex-simple sentence pairs and their corresponding human ratings \citep{xu-etal-2015-problems, xu2016optimizing, maddela-etal-2023-lens}.
Formally, given a source sentence $c$, a target simplification $t$, and optionally, a set of reference (human) ratings $r_1(c, t), ... , r_n(c, t)$, the task of evaluating text simplification is to compute a real-valued score $q(c, t)$.
The strength of the evaluation is usually measured by the correlation of $q$ with the (aggregated) reference ratings.
Evaluation methods are considered reliable if they have high correlation with human raters.
Hence, it is very important to consider the quality of source complex sentences $c$ and ratings $r$ in a benchmark for reliable evaluation.

We investigate three such benchmarks commonly used for text simplification: Simplicity-DA \citep{alva-manchego-etal-2021-un}, Newsela-Likert \citep{maddela-etal-2021-controllable}, and SimpEval2022 \citep{maddela-etal-2023-lens}.
The Simplicity-DA dataset, also referred to as WikiDA, consists of 600 simplifications of 100 complex sentences collected from Wikipedia. 
The simplifications are sourced from the TurkCorpus test set \citep{xu2016optimizing} and are produced by six older systems: PBMT-R, Hybrid, SBMT-SARI, Dress-Ls, DMASS-DCSS, and ACCESS.
These are rated by humans on three separate criteria (fluency, meaning, and simplicity) on a scale of 0-100.
The Newsela-Likert dataset contains 500 simplifications from 100 complex sentences sourced from the Newsela dataset \citep{xu-etal-2015-problems}, with one human simplification and four system simplifications.
The systems are BERT-Initialized Transformer \citep{jiang2020neural}, EditNTS \citep{dong-etal-2019-editnts}, LSTM \citep{zhang2017sentence}, and Hybrid-NG \citep{narayan2014hybrid}.
Newsela-Likert also contains human ratings on fluency, meaning, and simplicity, but on a five point scale.
Lastly, SimpEval2022 contains 360 simplifications from 60 complex sentences, collected from Wikipedia and simplified by humans and previously SOTA models including GPT-3.5 \citep{brown2020language} and T5 \citep{raffel2020exploring}.
SimpEval2022 is also scored on a 0-100 scale.

In our analysis, we find two major problems across existing datasets.
First, many of the sentence pairs are overly simplistic or incoherent; examples are shown in \autoref{tab:old_data_samples}.
As such, they might not reflect modern systems or accurately assess the model's simplification capabilities.
Second, we find lower agreement among human raters, indicating subjectivity of the annotation task, possible underspecification of the task instructions, and issues with annotation collection \citep{nowak2010reliable, aroyo2014three, hsueh2009data}.
Coupled with the relatively small sample size for each sentence pair, it is difficult to construe these ratings as a reliable gold standard.

\subsection{Coherence and Difficulty Gaps in Existing Datasets}
\label{sec:coherence_difficulty_existing_datasets}

We first investigate the sentence pairs in existing datasets and find many that are not reflective of the output of modern simplification systems such as LLMs.
This is particularly prevalent in older datasets, where simplifications were often generated by dated systems as opposed to written by hand.
A few example simplifications are shown in \autoref{tab:old_data_samples}. Many low-quality simplifications in existing datasets \citep{alva-manchego-etal-2021-un} reflect failure modes that modern LLMs rarely exhibit, like grammatical errors and disfluencies \citep{reinhart2025llms}. 
This results in a mismatch between the kinds of "bad" simplifications present in these datasets and the kinds of errors made by contemporary systems, highlighting the need for new, more representative examples of low-quality simplifications.

\input{tables_old_data_samples}

Secondly, the overall difficulty level of existing datasets is relatively low \citep{maddela-etal-2021-controllable, maddela-etal-2023-lens} and does not reflect the complexity of potential real-world applications, such as those in medicine or law. 
Sentences that are considered difficult in existing datasets often require moderate in-domain knowledge to understand, but the vocabulary itself is not overtly technical nor the sentence structure grammatically complex.
While this is not inherently problematic, as simplification can be evaluated even on simple inputs, the lack of linguistic and semantic complexity means these datasets fail to adequately challenge modern LLMs. 
As LLMs can now reliably produce fluent and coherent simplifications (\S\ref{sec:synthetic_dataset_construction}), such data does little to test their capabilities in preserving nuance, accurately simplifying domain-specific content, or avoiding hallucinations \citep{devaraj-etal-2022-evaluating}, falling short of providing a robust evaluation for current systems.

\subsection{Annotator Disagreement}
\label{sec:annotator_disagreement}
\input{tables_dataset_iaa}

Previous literature has shown that human annotators are not always in agreement across many other NLP tasks \citep{castilho-2020-page}.
In our preliminary experiments, we find that this pattern holds in existing text simplification datasets.
We compute inter-annotator agreement across the Newsela and SimpEval2022 datasets and report the $ICC(2, 1)$ (Intraclass Correlation) scores in \autoref{tab:dataset_iaa}.
ICC is often used to compute consistency across raters on non-categorical data \citep{hackl2023gpt}.
We find that individual raters are on average unreliable, which brings the relevance of their averages into question.
It may still be possible to find a more accurate rating through larger sample sizes, but this is often prohibitively expensive.

Moreover, text simplification a relatively subjective task and individual performance heavily depends on the quality of instructions and examples provided. 
What is “simple” can vary significantly depending on the educational background, the familiarity of the topic, and prior knowledge \citep{aroyo2015truth, liu2018inverse}.
Previous literature does not report the educational background of its annotators, nor does it attempt to control for variability across annotator profiles \citep{snow2008cheap}.
A correct-sounding, but semantically incorrect, simplification of a technical statement may not be correctly identified by a rater unfamiliar with the subject.
Secondly, it is important for the raters to have a strong set of guidelines in performing the ratings. Previous work often leaves this up to interpretation, allowing the rater to decide on their own definition of “simplicity” \citep{alva-manchego-etal-2021-un}.
This lack of transparency and standardization adds another layer of uncertainty to the reliability of the gold standard ratings in the datasets.

\section{\benchmarkname: A Synthetic Benchmark for Evaluating Text Simplification}
\label{sec:synthetic_benchmark}

To address the challenges in existing benchmarks and corresponding human ratings,
we introduce a new synthetic benchmark, or  \benchmarkname, using Qwen 2.5 72B Instruct (\S\ref{sec:synthetic_dataset_construction}).
We assess the quality of our benchmark on the basis of two criteria (\S\ref{sec:model_size_human_pilot}).
First, we measure the agreement among human annotators for simplifications in a pilot study.
Second, we report the performance of models of different sizes, based on our human pilot.

\subsection{Constructing \benchmarkname}
\label{sec:synthetic_dataset_construction}
\input{tables_synthetic_samples}

We construct our dataset by combining human-written and synthetic complex sentences. 
We use all 60 news article sentences from SimpEval2022, which are more challenging than other datasets, with an average FKGL of 18.29 compared to 8.79 for Newsela-Likert and 10.61 for Simplicity-DA. 
Additionally, we generate 200 synthetic sentences using Qwen 2.5 72B Instruct \citep{qwen2025qwen25technicalreport}. This was designed with 20 knowledge domains and 750 concept nouns  (full lists in \autoref{appendix:dataset_lists}), resulting in a diverse and challenging (average FKGL of $19.51$) test set. See example sentences in \autoref{tab:synthetic_samples} and the prompt used in \autoref{appendix:prompt_details}.

We collect simplifications of 260 complex sentences from four instruction-tuned Llama 3 models (1B, 3B, 8B, and 70B Instruct) using the same 2-shot prompt (details in \autoref{appendix:prompt_details}). With a total of 1040 simplified sentences whose relative quality we compare in \autoref{tab:simplified_sample}, our dataset is comparable or slightly larger than prior works.

In our benchmark, we use model size as a direct proxy for simplification quality, motivated by prior findings that LLMs generally produce higher-quality outputs (\S\ref{sec:intro}). We operationalize this intuition by selecting four LLaMA 3 models of increasing size (1B, 3B, 8B, and 70B), assuming larger models generate better simplifications. Rather than assigning numerical scores, we evaluate metrics based on their ability to correctly rank simplifications by model size using Spearman correlation, a method validated by our subsequent human study.

\subsection{Human Evaluation: Correlation with Model Size and Annotator Agreement}

\label{sec:model_size_human_pilot}

\begin{wrapfigure}{R}{0.3\textwidth}
\centering
\includegraphics[width=\linewidth]{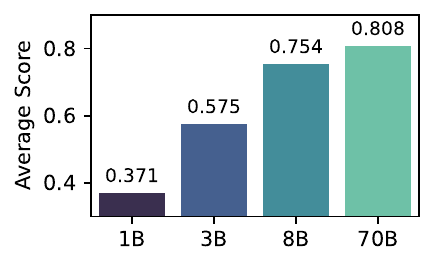}
\caption{Average human ratings of each model size, normalized to 0-1. We see that larger models consistently receive higher scores.
}
\label{fig:average_human_rating}
\end{wrapfigure}

To validate our assumption that larger models produce better simplifications, we conduct a human evaluation on 20 randomly selected complex sentences from \benchmarkname (10 each from SimpEval2022 and our synthetic dataset), paired with simplifications from four models (totaling 80 pairs). Three NLP-expert annotators rate simplification quality on a 5-point Likert scale using consistent instructions (Appendix \ref{app:judge-prompt}). As shown in \autoref{tab:dataset_iaa}, Inter-annotator agreement measured in ICC(2,1) is significantly higher than previous datasets, indicating our annotation process, with clear instructions, a well-defined rubric, and high-quality sentence pairs, promotes more consistent judgments.

Moreover, we find that human ratings strongly correlate with model size, as demonstrated in
\autoref{fig:average_human_rating}. 
Computing the Spearman rank correlation between human ratings and model size scores yields a correlation of 0.759. 
This result supports our assumption that model size is a meaningful proxy for simplification quality.
Given the limited scale of our human annotated dataset, we aim to further validate the effectiveness of \benchmarkname  at scale through automated evaluation using LLMs.

\section{Auto-Evaluation with \method on \benchmarkname}
\label{sec:evaluating_text_simplification}
\input{tables_simplified_sample}

We now introduce an \method framework for evaluating text simplification on \benchmarkname, given the difficulty and costs of attaining high quality ratings from human raters.
In our setup, we follow prior work in which multiple LLMs (as a jury of judges) are prompted to assess the quality of simplifications based on a shared rubric.

Drawing on previous work on text simplification that employs Likert scales for evaluation \citep{xu2016optimizing, alva-manchego-etal-2020-asset}, we prompt each LLM in our jury to produce a score from 1 to 5 for each simplification using the same prompt (see \autoref{appendix:prompt_details}).
We first construct a prompt $P = [I_{pre}; c; I_{mid}; t; I_{post}]$ by concatenation ($;$), where $c$ is the source sentence and $t$ is the target.
$I_{mid}$ and $I_{post}$ are primarily formatting, and $I_{pre}$ is a natural language instruction that elicits some score $s$ from the model.
We take these instructions and query multiple language models (LMs) $J_1, ..., J_n$ to collect a set of scores $S = {s_{1}, ..., s_{n}}$.
Using an aggregation function $f(S)$, we compute the final score $q = f(S)$.

We find that few-shot prompt with rationale generation performs best.
To enhance reproducibility, we use a lower temperature setting, though not zero, as we observe that small models occasionally produce outputs that are difficult to parse correctly.
In these cases, regenerating a response with a different seed addresses the problem.
Additionally, we consider a diverse range of instruction-tuned models to appear on our jury, from locally hosted ones to proprietary systems, allowing us to analyze how performance varies across different model architectures and sizes.
These models are the instruction-tuned versions of: Gemma 2 27B \citep{gemma_2024}, Qwen 2.5 32B \citep{qwen2025qwen25technicalreport}, Mixtral 8x7B \citep{jiang2024mixtralexperts}, Qwen 2.5 72B, Deepseek V3 \citep{deepseekai2025deepseekv3technicalreport}, Claude 3.7 Sonnet \citep{claude37sonnet}, and GPT-4o \citep{openai2024gpt4ocard}.
We normalize the collected scores to 0-1, and use the arithmetic mean of all models as our aggregation strategy.

\subsection{\method Ratings In Agreement}

Just as it is crucial that human raters produce a clear signal, our evaluation method must also be reliable, i.e., consistent among themselves.
We compute $ICC(2, 1)$ among the $7$ LLM judges, and find that they show strong agreement with each other (\autoref{tab:dataset_iaa} under LLM Unified).
Among the 3 SOTA models --- Deepseek V3, Claude 3.7 Sonnet, and GPT-4o) --- this is even higher at $0.754$.
The Spearman rank correlations between individual models can be found in \autoref{appendix:full_correlation_matrix}.

Simultaneously, many existing text simplification datasets \citep{alva-manchego-etal-2021-un, maddela-etal-2021-controllable} assess simplification quality across three dimensions: fluency, meaning preservation, and simplicity.
While this partitioning was useful in earlier evaluation settings, it may no longer be necessary or effective for modern LLMs.
In particular, SOTA LLMs rarely produce disfluent outputs, which reduces the informativeness of the fluency score.

To examine the utility of this traditional three-way partition, we compare it to a unified scoring approach in which LLMs are asked to provide a single overall rating of simplification quality.
Specifically, we prompt the LLM judges to independently rate fluency, meaning, and simplicity on a 1–5 Likert scale, and compute the inter-rater agreement across models using ICC(2,1).
We then compare these values to the ICC(2,1) score obtained when LLM judges give a single unified rating, as in our default setup.

As shown in \autoref{tab:dataset_iaa}, we find that the agreement among LLM judges is consistently higher when using unified scoring when compared to ratings for meaning and simplicity in the three-way partition.
This suggests that a unified rating not only simplifies the evaluation process but is also more reliable, offering a more stable signal for simplification quality in the context of LLM-generated outputs.

\subsection{LLM-as-a-Jury Ablation Studies}
\label{sec:llm_ablations}

\begin{figure}[h!]
\centering
\includegraphics[width=0.75\linewidth]{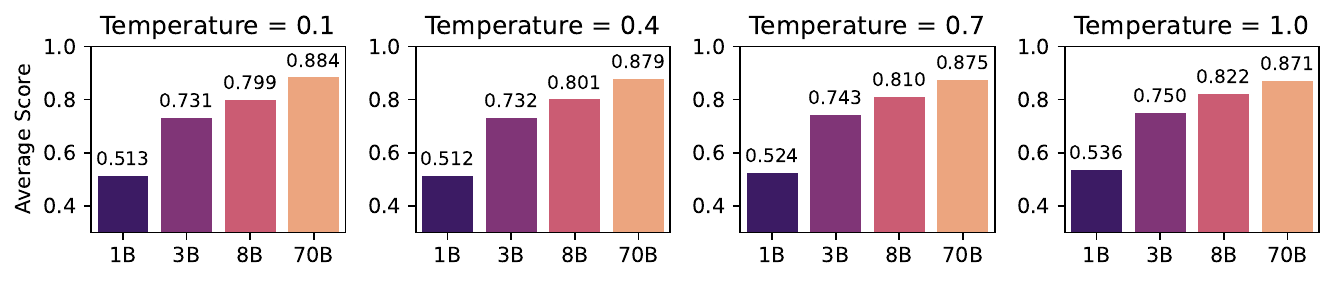}
\caption{Temperature ablation on \benchmarkname. Spearman rank correlations from left to right are 0.626, 0.622, 0.615, and 0.581 respectively.}
\label{fig:ablation_temp}
\end{figure}

In this section, we study the impact of various design choices on LLM judgment quality. 
As shown in \S\ref{sec:model_size_human_pilot}, model size is a proxy for simplification quality.
Therefore, we perform ablations to optimize correlation with model size, finding a strong setup for LLM judges that covers model selection, rationale generation, and few-shot prompts.

\begin{wrapfigure}{r}{0.6\textwidth}
\centering
\includegraphics[width=\linewidth]{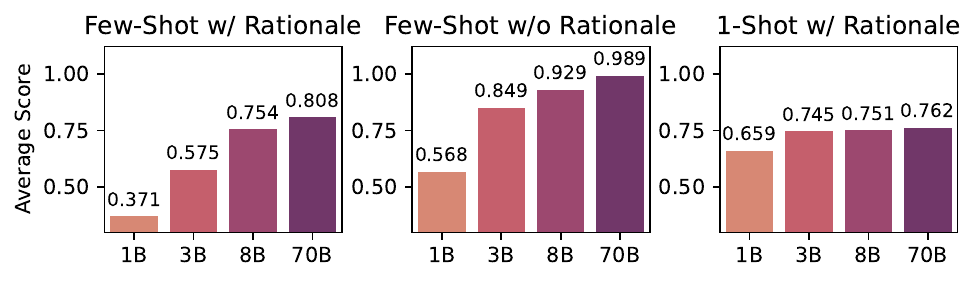}
\caption{Prompt format ablations on \benchmarkname. Spearman rank correlations from left to right are 0.626, 0.335, and 0.648 respectively.}
\label{fig:ablation_rationale}
\end{wrapfigure}

We perform our ablations with the following base configuration:
Our language model judges have a temperature of $0.1$ and a consistent prompt found in \autoref{appendix:prompt_details}.
Our base prompt is few-shot and asks the model to provide a rationale before answering.
Using this setup, we test all LLMs in our jury. Also, note that we do not test any Llama models as judges, as they are the simplifiers and may introduce bias through self-evaluation.
Apart from model selection, our other ablations use Qwen 2.5 72B Instruct as a judge.
The 1040 input sentence pairs are as described in \S\ref{sec:synthetic_dataset_construction}, and we compare the average scores of each of the four simplifier models.
We also compute their Spearman rank correlations with the model size score in \S\ref{sec:synthetic_dataset_construction}.

We find that temperature has minimal impact on judge performance.
A slight decrease in correlation is noticed, but this is likely due to randomness from increasing temperature.

\begin{wrapfigure}{r}{0.6\textwidth}
\centering
\vspace{-.25in}
\includegraphics[width=\linewidth]{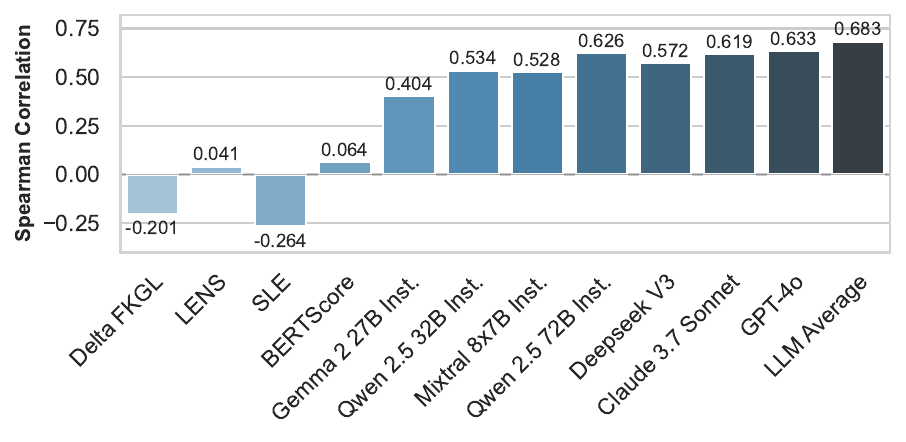}
\caption{Spearman correlations on \benchmarkname between existing metrics, LLMs, and LLM average with model size. See full correlation matrix in \autoref{appendix:full_correlation_matrix}.}
\label{fig:correlation_model_size}
\end{wrapfigure}

On the other hand, we find that few-shot prompts significantly outperforms 1-shot.
The latter prompt results in a large percentage of resulting scores being $4$ out of $5$, likely because all the simplifications are of decent quality.
Our few-shot examples may be encouraging the model to be stricter in its judgments, resulting in a flatter distribution of scores.
While the correlation with model size is similar, including the rationale depresses the average scores.
With rationale, the 70B simplifier model drops from a near-perfect average (without rationale) to $4.52$ out of $5$.
This leaves room for potentially stronger models --- such as the 405B variant of Llama 3 --- while preserving the model's ability to judge accurately.

Lastly, we perform an ablation on 7 judge models (\S\ref{sec:evaluating_text_simplification}) of various sizes and report results in \autoref{fig:correlation_model_size}.
We find that larger and closed-source models, on the right, tend to perform better than their smaller counterparts. Notably, the panel's overall judgment --- the average score --- performs markedly better than any individual model. While it is difficult to form an exact ranking, as correlation is a proxy, we find that model performance is consistent with general understandings of model output quality.

\subsection{Evaluating Existing Metrics and \method on \benchmarkname}
\label{sec:existing_methods}

First, we use our silver standard benchmark \benchmarkname to assess a set of widely used or new automatic evaluation metrics for text simplification: FKGL \citep{flesch1948new}, LENS \citep{maddela-etal-2023-lens}, BERTScore \citep{zhang2019bertscore}, and SLE \citep{cripwell2023simplicity}.
Note that we do not use systems that are significantly dependent on reference sentences, such as SARI, as our synthetic dataset does not include them.
Systems that are not totally dependent on references, such as LENS, are included by setting the reference to the simplification itself.

FKGL measures readability using sentence length and syllable count. We calculate Delta FKGL, the difference between simple and expert sentences, to assess relative simplicity.
We define it as follows: $\Delta FKGL = FKGL(c) - FKGL(t)$.
As a high FKGL represents a difficult sentence, $\Delta FKGL$ is high when the complex sentence is much harder than the simple sentence.
LENS uses an encoder transformer, RoBERTa-large, to encode sentences into vectors.
A trained feedforward neural network then predicts a single scalar score using these vectors as inputs.
Lastly, BERTScore, which is generally used for text similarity, has also been applied to measure text simplicity.
BERTScore generally compares word embeddings of the complex, reference, and simple sentences, while BLEU uses a formula that considers n-gram precision and sentence brevity.
Similar to FKGL, SLE is an absolute measure. However, it is instead computed by a fine-tuned LLM. We use $\Delta SLE$ as defined in \cite{cripwell2023simplicity} to measure a simplification's quality.

\begin{figure}[b!]
\centering
\includegraphics[width=0.95\textwidth]{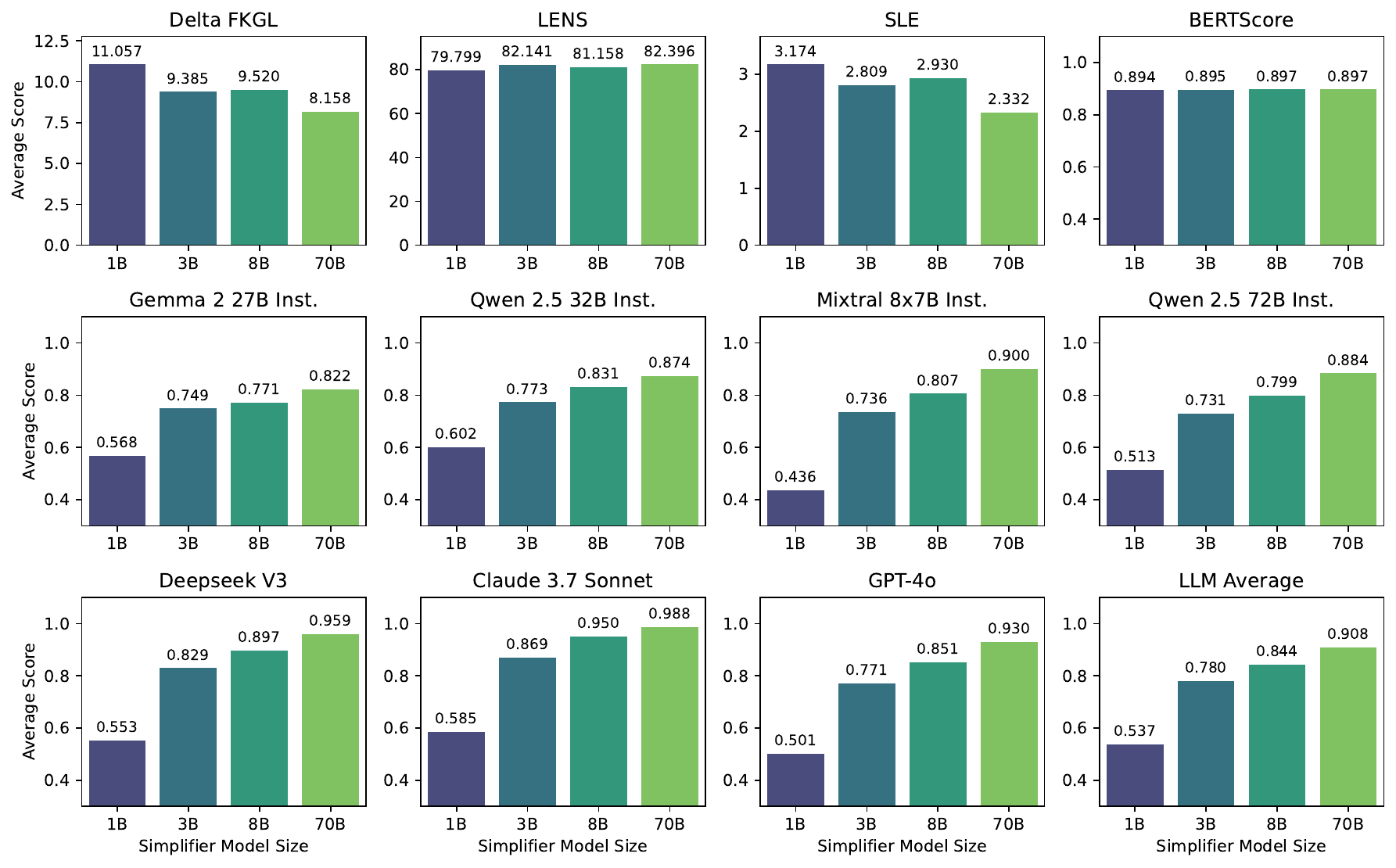}
\caption{Average scores of each simplifier (Llama 3) model. The top row contains previous metrics, and the bottom two rows contain normalized LLM judge scores (\S\ref{sec:evaluating_text_simplification}).}
\label{fig:sml_chart}
\end{figure}

We plot the average scores of each of the 4 simplifier models in \autoref{fig:sml_chart}.
None of the existing metrics show a strong correlation with the simplifier model's size.
On the contrary, FKGL and SLE exhibit an opposite correlation, where larger models receive lower simplification scores than smaller models.
While BERTScore and LENS seem to show some correlation with model sizes, average scores are very similar among the four models, making it difficult to differentiate between them.
Spearman correlations with model size are also low in \autoref{fig:correlation_model_size}.
This suggests that none of the existing metrics reliably capture the expected trend that previous works suggest and human raters identify.
Since existing evaluation metrics fail to reliably distinguish between simplifications generated by models of different sizes, we investigate our alternative approach that uses a panel of LLMs as evaluators.

As shown in \autoref{fig:sml_chart}, both individual LLMs and the aggregated panel consistently assign higher scores to simplifications produced by larger models.
This suggests that LLM judges are sensitive to subtle qualitative differences in simplification quality that scale with model size.
This indicates that existing simplification metrics --- FKGL, LENS, SLE, and BERTScore --- struggle to capture the key aspects necessary to evaluate the finer distinctions in LLM-generated simplifications.

\section{Can Existing Learnable Metrics Benefit from Synthetic Data?}
\label{sec:synthetic_data_training}

As mentioned in \S\ref{sec:coherence_difficulty_existing_datasets}, existing datasets generally do not reflect modern simplification systems.
Consequently, systems trained on lower quality data --- such as LENS \citep{maddela-etal-2023-lens}, SLE \citep{cripwell2023simplicity}, and REFeREE \cite{huang-kochmar-2024-referee} --- may learn information that is less applicable to modern simplification systems.
Additionally, the evaluation of multiple LLMs is often expensive, and it would be beneficial to have a smaller model that can perform faster, albeit less accurate, evaluation.
We train a small feedforward network with a similar architecture to LENS and SLE to explore these question.

Following existing work, we train a small neural network on the sentence embeddings of the complex and simplified sentences.
We collect a separate set of 400 synthetic sentences created in the same manner as \benchmarkname (\S\ref{sec:synthetic_dataset_construction}), each simplified by the 1B and 8B models; this pair of simplifiers produces a diverse range of ratings.
While previous work has used models such as RoBERTa, we compute embeddings with the SentenceTransformers library \citep{reimers-2019-sentence-bert}.
We use a newer embedding model, \texttt{all-mpnet-base-v2}, which is less than half the size of previous work.
We compute embeddings of length 768 for the complex and simple sentences, $E_{complex}$ and $E_{simple}$.
The input feature is constructed as $X = [E_{complex}; E_{simple}; E_{complex} - E_{simple}; E_{complex} \odot E_{simple}]$ where $\odot$ is the Hadamard product.
Our final network is a single-layer feedforward network with $64$ neurons --- much smaller than previous work --- to predict the score given to the input pair by the Qwen 2.5 72B Instruct judge.

We find that the resulting model has a correlation of $0.22$ with the model size score, substantially higher than previous metrics but much lower than the full model.
As a small model, this is a strong result which carries the implication that a potential issue facing existing metrics is the quality of their training data.

\section{Conclusion}
In this work, we address key limitations in existing text simplification benchmarks including low dataset quality and low annotator agreement. 
We address these limitations by introducing \benchmarkname, a synthetic benchmark designed to better reflect the capabilities of modern language models. 
Our benchmark combines complex inputs from both human-written and model-generated sources, and includes simplifications produced by models of varying sizes. 
Through a human study with high inter-annotator agreement and correlation with model size, we validate the reliability of our dataset. 
To scale evaluation and reduce annotation costs, we adopt a panel of LLM judges (\method), which produces scores that align strongly with model size and show higher consistency than existing metrics. 
We further demonstrate that learnable metrics can benefit from training on \method-labeled data, improving their ability to reflect true simplification quality. 
Overall, our findings suggest a practical and extensible approach for building reliable evaluation resources in tasks where high-quality annotations are limited or unreliable.

\input{colm2025_conference.bbl}
\clearpage
\appendix

\section{Related Work}
\label{sec:related_work}
Existing text simplification metrics broadly fall into two categories: static and learnable metrics. 
Traditional metrics were deterministic, 
and often depended on word or n-gram occurrence.
Examples of this include SARI \citep{xu2016optimizing} and BLEU \citep{papineni2002bleu}. 
Both SARI and BLEU consider n-gram similarity, and SARI further considers the importance of added and removed n-grams through the use of references.
Even earlier approaches included FKGL \citep{flesch1948new}, which is still commonly used. 
This computes text simplicity using a formula containing average syllables per word and words per sentence.
However, these metrics are not designed for the text simplification task, having been adapted from other fields such as machine translation; recent work has shown that this has limitations, such as negative correlations with simplicity on certain datasets \citep{sulem-etal-2018-bleu, tanprasert-kauchak-2021-flesch}.

More recently, work has been done on using LMs to measure text simplicity. While initially designed for semantic similarity, BERTScore \citep{zhang2019bertscore} has been used to measure some aspects of text simplification. More recent work, such as LENS \citep{maddela-etal-2023-lens}, REFeREE \citep{huang-kochmar-2024-referee}, and SLE \citep{cripwell2023simplicity}, have trained smaller models (such as RoBERTa) to predict scores. 
While they perform relatively well, they are also limited by the need to collect datasets with human ratings. 
SLE circumvents this by using a combination of Newsela data \cite{xu-etal-2015-problems} --- already labeled by difficulty --- and interpolation with FKGL, but this dataset is also constrained by the generalization of the former and performance of the latter.

Our work builds on language model inference techniques. 
We base our reasoning on chain-of-thought \citep{wei2022chain}, adapted to a classification task, and use few-shot learning \citep{brown2020language}; in particular, one-shot learning greatly improves performance. 
Lastly, we use models as evaluators, which have previously shown performance competitive with, and in some cases superior to, human judgment \citep{bohnet2022attributed}. 
Additionally, pre-trained models are able to generalize better than their fine-tuned counterparts \citep{huang2024empiricalstudyllmasajudgellm}.
However, one main drawback is that these models tend to prefer their own outputs \citep{panickssery2024llm}.
To counteract this, we use juries as proposed by \citet{verga2024replacing}  to improve performance while decreasing hardware requirements and costs.
We additionally take steps to ensure that models judge neither their own outputs nor the outputs of other models in their family.

\clearpage
\section{Full Correlation Matrix}
\label{appendix:full_correlation_matrix}
To compute the full correlation matrix, each of the existing metrics and LLM judges score the 1040 data pairs in \benchmarkname, and we compute the Spearman rank correlation. We additionally add the average LLM score and the model size score. The various LLMs have high correlations with each other as well as with the model size score. On the other hand, existing metrics do not correlate well with either.
\begin{figure}[h!]
 \centering
 \includegraphics[width=\linewidth]{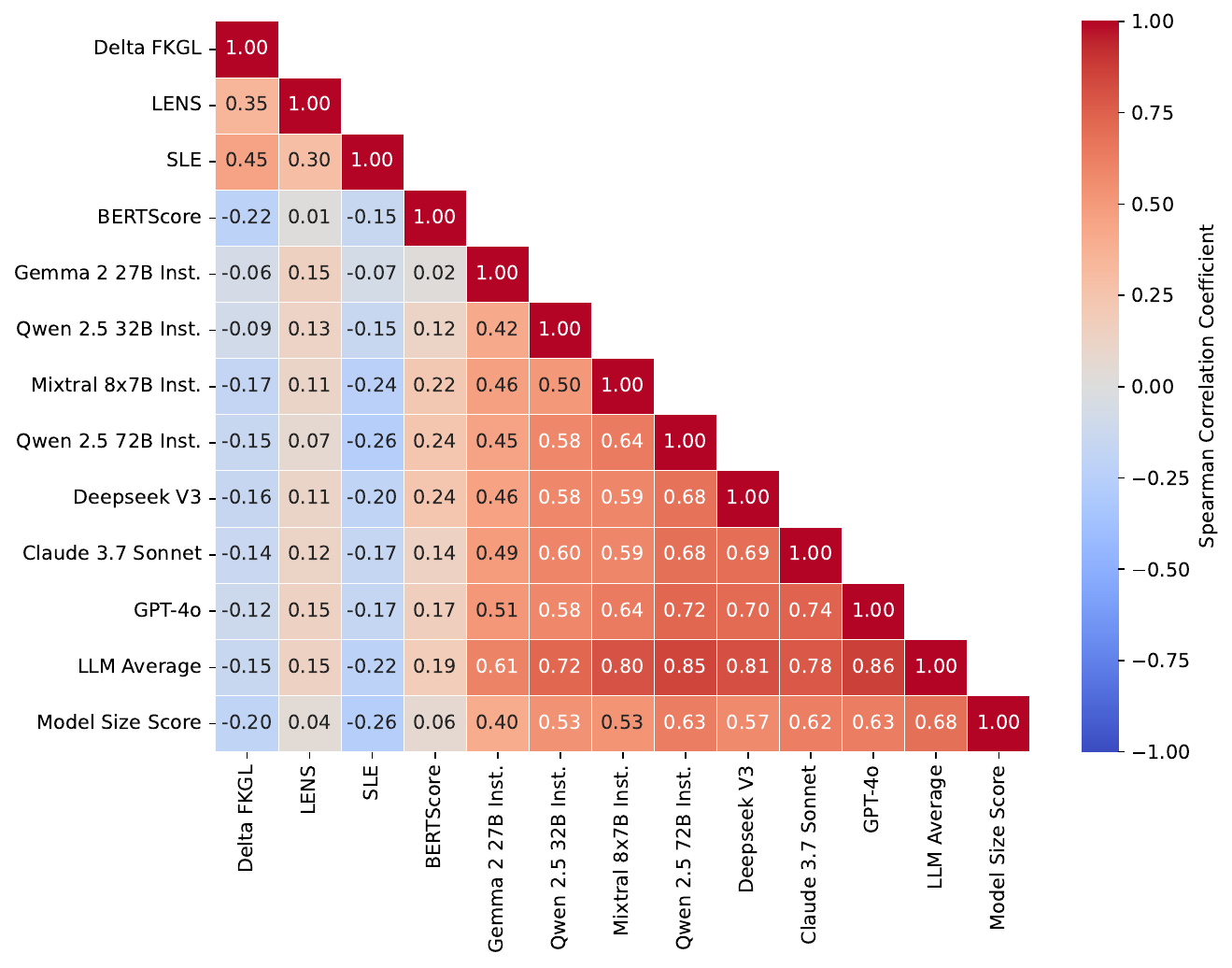}
 \caption{The full correlation matrix between existing metrics, LLM judges, their average, and the model size score from \S\ref{sec:existing_methods}.}
 \label{fig:correlation_matrix_large}
\end{figure}

\section{Prompt Details}
\label{appendix:prompt_details}
We use a chat format for all of our inference tasks.
Some prompts have a system role message containing instructions.
For models that don't support a system role, we simply prepend it to the first message.
The prompts provided are in the ChatML format, but are replaced automatically with whichever format the model defines for vLLM.
We additionally have prompts for our various ablations; due to length, however, these can be found in our GitHub repository.

\subsection{Synthetic Data Generation}
We begin our data generation by asking for complex sentences about a subject in a domain of knowledge (\autoref{appendix:dataset_lists}). This is a 1-shot prompt, with no system role.

\begin{verbatim}
<|im_start|>user
Please provide a technically difficult sentence about physical education. The
sentence should be concise but specific; instead of overcomplicating, try to come
up with something that would be found in a technical report or paper. If you wish,
you may consider the following subject, which may or may not be related: 'school'.
<|im_end|>

<|im_start|>assistant
The PE department's allocation of resources must ensure adequate supervision of
students engaging in high-impact aerobic activities.
<|im_end|>

<|im_start|>user
Please provide a technically difficult sentence about {subject}. The
sentence should be concise but specific; instead of overcomplicating, try to come
up with something that would be found in a technical report or paper. If you wish,
you may consider the following subject, which may or may not be related:
'{seed_noun}'.
<|im_end|>
\end{verbatim}

We then ask the various Llama models to simplify the resulting sentences with the following prompt. We use four instruction-tuned models from the Llama 3 family: Llama 3.2 1B Instruct, Llama 3.2 3B Instruct, Llama 3.1 8B Instruct, and Llama 3.3 70B Instruct.
We found that a 2-shot prompt produced more reliable outputs, particularly on smaller models.

\begin{verbatim}
<|im_start|>system
As an expert educator, your task is to simplify the provided sentences into a form
that elementary school children would find easier to understand. In your
simplification, try to make the sentence easier to understand while retaining the
original meaning. Replace technical terms with easier words, particularly those
that are more common. If needed, break the original sentence into multiple
sentences to simplify sentence structure - it is perfectly fine to extend the
length of the output if it makes it easier to understand. Your output will be
judged on 3 aspects: fluency (is it grammatically correct; does it flow well?),
meaning (does it preserve the original meaning and nuance?), and most importantly
simplicity (is it easier to read and understand?). Output only the simplified
sentence(s) on a single line.
<|im_end|>

<|im_start|>user
Please simplify the following sentence:\n
The cumulative effect of instructors' accrued sick leave and retirement benefits
must be reconciled with the department's annual expenditure on personnel-related
overheads.
<|im_end|>

<|im_start|>assistant
The money used for teacher's sick time and retirement pay has to match up with the
money the department spends on all its workers every year.
<|im_end|>

<|im_start|>user
Please simplify the following sentence:\n
The prolonged exposure to screens can lead to a significant alteration in the
dopaminergic reward system, potentially affecting cognitive and behavioral
outcomes in adolescents.
<|im_end|>

<|im_start|>assistant
Looking at screens for a long time can change how your brain feels happy and
rewards you. This might affect how young people think and behave.
<|im_end|>

<|im_start|>user
Please simplify the following sentence:\n
{sentence}
<|im_end|>
\end{verbatim}

\subsection{LLM Judge Prompt}
\label{app:judge-prompt}
We provide our final LLM judge prompt. This is a few-shot prompt with a system message.

\begin{verbatim}
<|im_start|>system
You are an expert linguist and educator. Your task is to judge how well a system
has simplified a sentence. To do this, you will be given a base sentence, which 
will be more complex, and a new version of that sentence. In your judgement, 
consider text simplicity (how hard is it for children to read?), semantic 
similarity (does it mean the same thing, and convey the same nuance?), and fluency 
(the simplification should be grammatically correct). First, concisely consider 
all three aspects. Note that a sentence can do well in some aspects and badly in 
others - use your judgement to decide which is more important for that sentence. 
If the sentences are identical, your answer should be around a [3] - but also 
consider whether its possible to simplify it further. At the end, provide an 
integer score from [1] to [5], where [5] represents a high quality simplification, 
in square brackets. Provided below is a general rubric:\n
[5]: The sentence is fully simplified, entirely fluent, and preserves the core 
meaning of the original sentence.\n
[4]: The sentence is somewhat simpler, mostly fluent, and the meaning is close to 
the original sentence.\n
[3]: The sentence is simpler, somewhat fluent, and the meaning is similar to the 
original sentence.\n
[2]: The sentence is equivalently simple, still has some fluency, but the meaning 
is lost.\n
[1]: The sentence is completely unreadable.
<|im_end|>

<|im_start|>user
Base Sentence: It will then dislodge itself and sink back to the river bed in 
order to digest its food and wait for its next meal.\n
New Sentence: It will then move away from the river bed and sink back to the 
bottom to digest its food.
<|im_end|>

<|im_start|>assistant
The sentence reads completely fluently. The original meaning of the sentence - 
something unattaching and sinking to the bottom - is altered, as the subject 
"moves away" from the river bed in the simple sentence as opposed to moving 
towards it. It is somewhat simpler than the original, such as by using "move away" 
instead of "dislodge". Answer: [3]
<|im_end|>

<|im_start|>user
Base Sentence: As a result, although many mosques will not enforce violations, 
both men and women when attending a mosque must adhere to these guidelines.\n
New Sentence: Due to this, a lot of mosques don't enforce these rules but both men 
and women should follow them.
<|im_end|>

<|im_start|>assistant
Although the new sentence has an error in fluency (missing a comma after "these 
rules"), it has the same meaning as the original. The sentence also uses simpler 
vocabulary (such as by using "follow" instead of "adhere") and sentence 
structure, notwithstanding the grammatical error. Answer: [4]
<|im_end|>

<|im_start|>user
Base Sentence: A Wikipedia gadget is a JavaScript and/or a CSS snippet that can be 
enabled simply by checking an option in your Wikipedia preferences.\n
New Sentence: A a gadget javascript a is and / checking wikipedia an sinppet that 
can be enabled simply by or css option in your wikipedia preferences.
<|im_end>

<|im_start|>assistant
The new sentence is completely unreadable. Answer: [1]
<|im_end|>

<|im_start|>user
Base Sentence: Despite what many consider obvious, that which we pursue with 
greatest fervor often eludes our grasp not for lack of effort, but precisely 
because our desperate clutching drives away the very thing we seek.\n
New Sentence: Despite what many consider obvious, that which we pursue with 
greatest fervor often eludes our grasp not for lack of effort, but precisely 
because our desperate clutching drives away the very thing we seek.
<|im_end|>

<|im_start|>assistant
The new sentence is equivalent in simplicity, fluency, and meaning to the 
original. However, there are many ways to improve the sentence's readability. 
Answer: [3]
<|im_end|>

<|im_start|>user
Base Sentence: The dog ran across the green field to catch the ball.\n
New Sentence: The dog ran across the green field to catch the ball.
<|im_end|>

<|im_start|>assistant
The new sentence is equivalent in simplicity, fluency, and meaning to the 
original. However, it would be difficult to simplify this sentence further, as it 
is already very straightforward. Answer: [4]
<|im_end|>

<|im_start|>user
Base Sentence: {base_sentence}\n
New Sentence: {new_sentence}
<|im_end|>
\end{verbatim}
\section{Synthetic Dataset Generation Lists}
\label{appendix:dataset_lists}

\subsection{Domains of Knowledge}
We select $20$ domains of knowledge to ensure our datasets cover a wide range of potential sentences. These are:

\begin{center}
\small
\begin{tabular}{@{} l l l l @{}}
\hline
\multicolumn{4}{c}{Domains of Knowledge} \\
\hline

architecture & chemical engineering & physics & chemistry \\
art & nursing & music & biology \\
mathematics & philosophy & theater & anthropology \\
english & education & accounting & history \\
computer science & political science & economics & psychology \\
\end{tabular}
\end{center}

\subsection{Concept Nouns}

Our concept nouns were collected by filtering a list of common English nouns by removing words with common verb or adjective forms. The full list of $739$ filtered concept nouns is as follows.

\small
\begin{longtable}{@{} l l l l l @{}}
\hline
\multicolumn{5}{c}{Concept Nouns} \\
\hline
people & history & art & world & information \\
map & family & government & system & computer \\
meat & year & music & person & method \\
data & food & theory & law & bird \\
literature & problem & software & knowledge & ability \\
economics & internet & television & science & library \\
fact & product & idea & temperature & investment \\
society & activity & story & industry & thing \\
oven & community & definition & safety & quality \\
development & language & management & player & variety \\
video & country & exam & movie & organization \\
equipment & physics & analysis & policy & series \\
direction & strategy & technology & army & camera \\
freedom & environment & child & month & truth \\
university & writing & article & department & difference \\
goal & audience & growth & income & marriage \\
user & combination & failure & medicine & philosophy \\
teacher & communication & chemistry & disease & energy \\
nation & road & soup & location & success \\
apartment & education & painting & politics & decision \\
event & property & student & wood & competition \\
distribution & entertainment & office & population & president \\
unit & category & cigarette & context & introduction \\
opportunity & performance & driver & flight & length \\
magazine & newspaper & relationship & cell & dealer \\
finding & lake & member & phone & scene \\
association & concept & customer & discussion & housing \\
inflation & insurance & woman & effort & expression \\
importance & opinion & payment & reality & responsibility \\
situation & skill & wealth & application & city \\
county & depth & estate & foundation & grandmother \\
perspective & photo & recipe & studio & topic \\
collection & depression & imagination & resource & agency \\
college & connection & criticism & debt & description \\
patience & secretary & solution & administration & director \\
personality & psychology & recommendation & selection & alcohol \\
complaint & contract & highway & loss & membership \\
possession & preparation & steak & union & agreement \\
cancer & currency & employment & engineering & interaction \\
mixture & region & republic & tradition & virus \\
actor & classroom & delivery & device & difficulty \\
drama & election & engine & football & guidance \\
hotel & owner & protection & suggestion & variation \\
anxiety & atmosphere & awareness & bath & bread \\
candidate & comparison & confusion & construction & elevator \\
emotion & employee & employer & guest & leadership \\
mall & manager & operation & recording & sample \\
transportation & charity & cousin & disaster & editor \\
efficiency & excitement & guitar & homework & leader \\
outcome & presentation & promotion & refrigerator & resolution \\
revenue & session & singer & tennis & basket \\
bonus & cabinet & childhood & church & clothes \\
dinner & drawing & initiative & judgment & lab \\
measurement & mud & poetry & police & possibility \\
procedure & queen & relation & restaurant & satisfaction \\
sector & signature & significance & song & tooth \\
town & vehicle & volume & wife & accident \\
airport & arrival & baseball & chapter & committee \\
conversation & database & enthusiasm & explanation & farmer \\
gate & girl & hall & historian & hospital \\
injury & instruction & manufacturer & meal & perception \\
pie & poem & proposal & reception & replacement \\
revolution & river & son & speech & village \\
winner & worker & writer & assistance & buyer \\
chest & chocolate & conclusion & contribution & cookie \\
courage & desk & drawer & establishment & examination \\
garbage & grocery & improvement & independence & insect \\
inspection & inspector & king & ladder & penalty \\
piano & potato & profession & professor & quantity \\
requirement & salad & sister & supermarket & weakness \\
wedding & ambition & analyst & apple & assignment \\
assistant & bathroom & bedroom & celebration & championship \\
cheek & client & consequence & departure & diamond \\
dirt & fortune & friendship & gene & girlfriend \\
hat & lady & negotiation & obligation & passenger \\
pizza & platform & poet & pollution & recognition \\
reputation & shirt & speaker & stranger & surgery \\
tale & trainer & uncle & youth & film \\
water & money & example & business & study \\
game & field & fish & experience & job \\
book & economy & body & market & state \\
radio & company & card & list & group \\
force & key & training & school & research \\
service & web & boss & sport & house \\
page & soil & oil & picture & garden \\
site & exercise & image & case & coast \\   
action & boat & result & section & building \\
mouse & cash & class & store & tax \\
space & rule & model & source & earth \\
program & chicken & purpose & question & rock \\
salt & birth & car & dog & object \\
scale & sun & war & bank & craft \\
bus & eye & fire & box & frame \\
step & cycle & metal & room & screen \\
structure & ball & discipline & gift & machine \\
tool & career & culture & pot & sign \\
table & task & egg & ice & network \\
star & challenge & brush & plant & wing \\
brain & button & foot & wall & distance \\
pair & savings & staff & sugar & target \\
animal & author & budget & file & ground \\
lesson & officer & sky & stage & stick \\
title & bowl & bridge & campaign & character \\
club & evidence & fan & letter & novel \\
park & quarter & baby & dish & fruit \\
glass & muscle & strength & vegetable & chart \\
gear & kitchen & land & log & mother \\
relative & street & tree & bench & commission \\
path & project & sea & ticket & confidence \\
daughter & doctor & dot & duty & essay \\
father & milk & pipe & seat & stable \\
storm & substance & team & bat & beach \\
chain & consideration & cream & crew & gold \\
interview & kid & mission & shop & suit \\
window & agent & band & block & bone \\
calendar & cap & coat & contest & court \\
cup & district & door & finger & garage \\
hole & hook & layer & lecture & meeting \\
nose & rice & telephone & airline & bag \\
battle & bed & cake & designer & dimension \\
dress & emergency & extension & farm & horror \\
horse & husband & mountain & nail & noise \\
occasion & package & patient & phrase & sand \\
sentence & stomach & string & tourist & towel \\
vacation & wheel & wine & arm & associate \\
border & branch & brother & coach & document \\
expert & floor & god & iron & judge \\
knife & landscape & league & parent & pin \\
pool & pound & salary & shelter & shoe \\
tank & bell & bike & boy & brick \\
chair & closet & clue & collar & conference \\
devil & glove & jacket & monitor & mortgage \\
nurse & peak & plane & reward & sandwich \\
yard & bicycle & bottle & cable & candle \\
clerk & cloud & concert & counter & flower \\
grandfather & lawyer & mirror & pension & plate \\
ruin & ship & skirt & snow & specialist \\
trash & anger & award & boot & bug \\
camp & candy & carpet & cat & champion \\
clock & cow & engineer & entrance & grass \\
incident & island & jury & leg & lip \\
motor & nerve & passage & pen & priest \\
prize & resident & resort & ring & roof \\
rope & scheme & script & sock & station \\
toe & tower & truck & witness & human \\
individual & guard & watch & official & press \\
spring & objective & chemical & dump & conflict \\
mobile & train & bear & representative & \\
\end{longtable}

\section{Existing Dataset Samples}
\label{appendix:existing_samples_extended}
\normalfont
We provide randomly selected additional samples of complex-simple sentence pairs from both previous works and \benchmarkname.

\begin{table}[h!]
\centering
\begin{tabular}{p{2cm}p{5cm}p{5cm}}
\toprule
\textbf{Source} & \textbf{Complex Sentence} & \textbf{Simplified Sentence} \\
\midrule

Simplicity-DA & These works he produced and published himself, whilst his much larger woodcuts were mostly commissioned work. & These works he made and published himself and his much larger woodcuts were written work. \\[6pt]
\midrule
Simplicity-DA & The SAT Reasoning Test (formerly Scholastic Aptitude Test and Scholastic Assessment Test) is a standardized test for college admissions in the United States. & The SAT Reasoning Test (used to be called Scholastic Aptitude Test and Scholastic Assessment Test) is a test for college admissions in the United States. \\[6pt]
\midrule
Newsela-Likert & president barack obama understands that if he were to proclaim a goal of definitively eliminating isis in the short term, he would fail. & president barack obama knows that he can not promise to destroy the islamic state quickly. he would fail. \\[6pt]
\midrule
Newsela-Likert & he could not move or talk and he looked like he was sleeping. & he was hurt badly and could not move or even open his eyes.\\[6pt]
\midrule
SimpEval2022 & Two sisters, Leah and Chantrelle, and their acquaintance Hosanna catch a steamboat from Saint Ann Parish in Jamaica to the United Kingdom, arriving in London's Notting Hill before moving to the Midlands. & Two sisters, Leah and Chantrelle, as well as their friend Hosanna catch a boat from Jamaica to London. Then, they will move to the midlands. \\[6pt]
\midrule
SimpEval2022 & Drone footage released by the Islamic State showed bombs being dropped on an ammunitions facility located in Deir ez-Zor, Syria, an area of contested control between the Islamic State and the Syrian government at the time. & The Islamic State and the Syrian government were fighting to control an area in Syria called Deir ez-Zor. During that time, the Islamic State released videos that showed bombs being dropped on a weapons storage facility in that area. \\[6pt]
\midrule
\benchmarkname & The historical development of elevator technology in urban high-rise buildings significantly impacted the architectural design and social stratification of cities in the early 20th century. & In the early 1900s, buildings with high ceilings became common in cities. This made the people who lived in high buildings feel like they were better than those in lower buildings, and it changed the way people's homes were designed. \\[6pt]
\midrule
\benchmarkname & The significance of accurate variance analysis in financial reporting is paramount, as it directly impacts the reliability of financial statements and the decision-making processes of stakeholders. & The accuracy of financial reports is very important. Without it, people don't trust the information and can't make smart decisions about the company's business. \\[6pt]
\bottomrule
\end{tabular}
\caption{Additional sentence pairs}
\label{tab:existing_samples_appendix}
\end{table}

\end{document}

%% file: tables_old_data_samples.tex
\begin{table}
\centering
\begin{tabular}{p{2.2cm}p{5.5cm}p{4.9cm}}
\toprule
\textbf{Source} & \textbf{Complex Sentence} & \textbf{Simplified Sentence} \\
\midrule
Simplicity-DA & \small For example, the stylebook of the Associated Press is updated annually. & \small , the stylebook is updated. \\[6pt]
\midrule
Newsela-Likert & \small companies want to drill for oil in the park. & \small companies want to drill for oil. \\[6pt]
\midrule
SimpEval2022 & \small On the fifth day of flight, November 20, 2022, at 19:09 UTC, the Orion spacecraft entered the Lunar sphere of influence, thus the Moon's gravitational force became stronger than Earth's relative to the spacecraft. & \small On the fifth day of flight, November 20, 2022, at 19:09 UTC, the Orion spacecraft entered the Lunar sphere of influence, so the Moon's gravitational force became stronger than Earth's relative to the spacecraft. \\[6pt]
\midrule
\benchmarkname & \small The study's findings suggest a significant correlation between long-term cannabis use and alterations in neurocognitive function, particularly in attention and memory processes. & \small Using cannabis for a long time might change how your brain works, especially when it comes to paying attention and remembering things. \\[6pt]
\bottomrule
\end{tabular}
\caption{Example sentences from various datasets. The first complex sentence has a simplification that few modern systems would output as it is grammatically incorrect. Newsela-Likert and SimpEval2022 themselves are not challenging enough for modern LLMs to produce a useful simplification. Additional samples can be found in \autoref{appendix:existing_samples_extended}.
}
\label{tab:old_data_samples}
\vspace{-0.25in}
\end{table}

%% file: tables_dataset_iaa.tex
\begin{table}
  \footnotesize
  \centering
  \setlength{\tabcolsep}{3pt}
  \begin{tabular}{@{}lcccccccc@{}}
    \toprule
    \textbf{Dataset} & \textbf{Human} & \textbf{Human} & \textbf{Human} & \textbf{LLM} & \textbf{LLM} & \textbf{LLM} & \textbf{Human} & \textbf{LLM} \\
    & \textbf{Fluency} & \textbf{Meaning} & \textbf{Simplicity} & \textbf{Fluency} & \textbf{Meaning} & \textbf{Simplicity} & \textbf{Unified} & \textbf{Unified} \\
    \midrule
    Newsela              & 0.153 & 0.213 & 0.054 & 0.672 & 0.650 & 0.640 & --- & 0.657 \\
    SimpEval & --- & ---  & --- & --- & ---      & ---   & 0.228 & 0.320 \\
    \benchmarkname& ---  & ---   & ---  & ---   & ---  & ---  & 0.671* & 0.619 \\
    \bottomrule
  \end{tabular}
  \caption{Intraclass Correlation Coefficient, ICC(2, 1) for human and LLM ratings on three benchmarks. Human unified ratings(*) are on a subset of \benchmarkname (see \S\ref{sec:model_size_human_pilot}).}
  \label{tab:dataset_iaa}
\end{table}

%% file: tables_synthetic_samples.tex
\begin{table}
\centering
\begin{tabular}{p{2.4cm}p{10.6cm}}
\toprule
\textbf{Domain and \newline Concept Noun} & \textbf{Generated Sentence} \\
\midrule
\textbf{architecture} \newline grocery & \small The cantilevered roofline of the facility must accommodate 30-degree angular deviations in structural supports while maintaining a 3-inch minimum clearance from refrigerated storage units. \\
\midrule
\textbf{mathematics} \newline vacation & \small The optimization of vacation scheduling for a group can be modeled as a constraint satisfaction problem, where the objective function minimizes the total dissatisfaction across all participants, subject to constraints on available dates and group size. \\
\midrule
\textbf{anthropology} \newline alcohol & \small Ethnographic studies reveal that the ritualized consumption of alcohol in social gatherings serves as a significant mediator of interpersonal relationships and community cohesion in diverse cultural contexts. \\
\bottomrule
\end{tabular}
\caption{Synthetic complex sentence samples from \benchmarkname.}
\vspace{-.25in}
\label{tab:synthetic_samples}
\end{table}

%% file: tables_simplified_sample.tex
\begin{table}
\centering
\begin{tabular}{p{1.2cm}p{11.8cm}}
\toprule
\textbf{Type} & \textbf{Sentence} \\
\midrule
\small Complex & \small The integration of advanced payment systems in chemical engineering plants necessitates a robust cybersecurity framework to protect sensitive operational data and financial transactions. \\
\midrule
\small Llama 3 1B & \small Using advanced payment systems in factories helps keep the important things safe. To keep this safe, strong computer security measures are needed. \\
\midrule
\small Llama 3 3B & \small To keep information about the plants and money safe, we need a strong system to protect it from hackers and keep track of all the payments. \\
\midrule
\small Llama 3 8B & \small Using computers in factories that make chemicals requires a strong way to keep the information and money safe from being hacked. \\
\midrule
\small Llama 3 70B & \small When we use new ways to pay for things in chemical plants, we need to make sure we have strong protection for our computers and money information so it doesn't get stolen or hurt. \\
\bottomrule
\end{tabular}
\caption{An example simplification of a synthetic sentence in our dataset by the four Llama 3 models. The simplification prompt can be found in \autoref{appendix:prompt_details}.
}
\label{tab:simplified_sample}
\vspace{-.25in}
\end{table}